# Using GPT-4 to Augment Unbalanced Data for Automatic Scoring

Luyang Fang[a†], Gyeong-Geon Lee[b†] and Xiaoming Zhai[b, c]*

[a]*Department of Statistics, Franklin College of Arts and Sciences, University of Georgia, 310 Herty Drive, Athens, GA 30602, United States*

[b] *AI4STEM Education Center, Mary Frances Early College of Education, University of Georgia, 110 Carlton Street, Athens, GA 30605, United States*

[c] *Department of Mathematics, Science, and Social Studies Education, Mary Frances Early College of Education, University of Georgia, 110 Carlton Street, Athens, GA 30605, United States*

[†] These authors contributed equally to this work.

*corresponding author: Xiaoming.Zhai@uga.edu

**Abstract**

Machine learning-based automatic scoring faces challenges with unbalanced student responses across scoring categories. To address this, we introduce a novel text data augmentation framework leveraging GPT-4, a generative large language model, specifically tailored for unbalanced datasets in automatic scoring. Our experimental dataset comprised student written responses to four science items. We crafted prompts for GPT-4 to generate responses, especially for minority scoring classes, enhancing the data set. We then finetuned DistillBERT for automatic scoring based on the augmented and original datasets. Model performance was assessed using accuracy, precision, recall, and F1 metrics. Our findings revealed that incorporating GPT-4-augmented data remarkedly improved model performance, particularly for precision and F1 scores. Interestingly, the extent of improvement varied depending on the specific dataset and the proportion of augmented data used. Notably, we found that a varying amount of augmented data (20%-40%) was needed to obtain stable improvement for automatic scoring. Comparisons with models trained on additional student-written responses suggest that GPT-4 augmented models match those trained with student data. This research underscores the potential and effectiveness of data augmentation techniques utilizing generative large language models like GPT-4 in addressing unbalanced datasets within automated assessment.

# 1 Introduction

One of the most prominent innovations in recent educational assessment and measurement studies is the automation of scoring using machine learning -- a subset of artificial intelligence (AI) (Zhai et al. 2020). These applications are particularly useful in subjects such as science education, as science learning necessitates engaging students in knowledge uses and solution design, which need innovative performance-based assessments (NRC, 2014), particularly those that enable writing constructed responses to explain science phenomena. However, it has long been noted that scoring massive constructed responses was time- and cost-consuming. Although automated scoring methods are desired (Page 1966), it was not until the 2010s, when machine learning techniques were improved, researchers started to develop scoring models to assess students' written responses. Since then, machine learning technologies have evolved significantly, and researchers have utilized various machine learning techniques (e.g., decision tree, regression, ensemble, BERT) for scoring students' written explanations, arguments, description of models, etc. (Liu et al. 2023; Wu et al. 2023; Zhai et al. 2020; Jescovitch et al. 2021; Lee et al. 2023).

While the previous studies reported promising results, the scoring accuracies were found at a varying degree of success (Zhai, Shi & Nehm, 2021). Among the many challenges for automatic scoring, data imbalance has emerged as a significant issue that can impact the scoring accuracy and lead to validity issues of assessment uses (Kaldaras et al. 2022). In educational settings, the distribution of student responses often exhibits a skewed pattern, with a majority of responses falling into specific scoring categories. Machine learning algorithms may struggle to accurately classify minority classes due to the insufficient numbers of training data in the specific categories. This imbalance can lead to decreased scoring accuracy, often at the expense of students belonging to the minority class. Therefore, researchers must secure as much data as possible to improve model performance, particularly for minority classes (Shorten and Khoshgoftaar 2019). However, this is not always feasible as collecting additional data needs significant effort, and sometimes can infringe the research design and ethics. Consequently, novel data augmentation methods are needed to enlarge the training dataset (Cochran et al. 2022; Wang et al. 2021; Zhang et al. 2015).

To deal with this challenge, we employed a generative large language model, GPT-4, to augment student written responses and improve the data balance. Specifically, we used prompt engineering to generate responses resembling student written responses to four science assessment tasks. We compared the parameters of scoring models developed based on the augmented dataset with those of the original dataset, as well as with the dataset with the same amount of additional student written responses. This study answers three research questions:

1. To what degree does GPT-4 augmented training data improve the scoring performance?

2. How efficient is GPT-4-based data augmentation in improving the scoring model performance?

3. How does the GPT-4-based data augmentation result resemble real student-written data?

**2 Literature Review**

*2.1 Data Augmentation*

Data augmentation is a critical technique in machine learning that aims to increase the size and diversity of training datasets to improve model performance (Zhang et al. 2015; Wang et al. 2021). While data augmentation is well-established in fields like computer vision, its application in NLP is relatively new (Shorten et al., 2021). This method is rapidly growing and recently gaining popularity among researchers focusing on text classification problems, where the availability of labeled data was often limited or unbalanced. Recent studies suggested that NLP data augmentation could enhance machine learning model performance for student achievement classification in e-learning (Cader 2018), at-risk student academic success prediction (Bell et al. 2021), or cheating detection (Zhou and Jiao 2022), showing the potential of data augmentation for education studies.

Earlier work of NLP data augmentation relies on manual hand-crafted feature extractors and classifiers, such as Bag-of-words and TFIDF (term-frequency inverse-document-frequency) (Jones, 1972) and bag-of-means on word embedding (Mikolov et al., 2013). These methods are often time-consuming to create feature extractors or classifiers for replacement words. To improve efficiency, researchers created systematic methods to automatically augment data diversity based on rules. For example, Zhang et al. (2015) developed synonyms words rules to replace English words from the mytheas dataset. Using the program WordNet, they determined synonym words or phrases by ranking their semantic closeness to the most frequently seen meaning. They then created a method to identify the optimal word count to replace words with a geometric distribution index. Using this method, Zhang et al. (2015) augmented multiple large datasets with two deep learning language models, ConvNets and Long-Short Term Memory. Comparing to the outcomes with traditional handcrafted methods, they found that the character-level ConvNet was most effective. Similar methods were also created, such as back translation (Yu et al., 2018), random swap (swapping the positions of two words in a sentence randomly), or random deletion (removing each word in a sentence randomly) (Wei & Zou, 2019). Even though the authors reported improved model performance, they acknowledged that new methods were needed as rule-based replacements may distort the original meaning and highly rely on other factors, such as the capacity of the language models.

More evolved data augmentation techniques seek to leverage the feature space of data. When features extracted from data are represented within a vector space, the coordinates of each data point in the feature space can be used to mathematically elicit new data points. For example, Chawla et al. (2002) developed the Synthetic Minority Over-sampling Technique (SMOTE) algorithm using the feature space of data. The SMOTE algorithm starts with a minority sample with k-nearest neighbors. Assuming that ($a,b$) is the minority sample and ($c,d$) is one of the $k$-nearest neighbors, a new sample can be synthesized as,

($e,f$) = ($a,b$) + rand(0-1) * ($c-a,d-b$)

SMOTE algorithm repeats this according to the hyperparameter $k$ and the size of augmenting data designated by the user. Chawla et al. (2002) reported that the SMOTE algorithm was successful in augmenting nine benchmark datasets. Mixup algorithm is also based on interpolation between two minority samples (Zhang et al. 2017). However, while SMOTE algorithm decides the interpolation factor from a uniform distribution between 0 and 1, MixUp algorithm randomly selects the factor from a Beta ($\alpha,\alpha$) distribution where the user can select hyperparameter $\alpha$. Guo et al. (2019) showed that MixUp algorithm effectively augments data for both image and text classification tasks and improves sentence classification model performance. However, there has been a technical bottleneck in using Mixup for its requirements of continuous inputs, which was remedied by more advanced algorithms (Feng et al. 2021). Besides, there are approaches combining the abovementioned techniques to expand the possibility of drawing more diversified instances (Bayer et al. 2022).

Advances in NLP data augmentation have recently been brought about by the generative language models (Antonious et al. 2017; Shorten et al. 2021; Bayer et al. 2022). For example, within the Generative Adversarial Network (GAN) framework, a 'generator' model is trained on the existing data to create candidates that resemble this data, while a 'discriminator' model determines if a given sample is from the original dataset or artificial to feed it back to enhance the 'generator' model performance (Goodfellow et al. 2014). Data generated by generative models could be used for training other models, which can reduce overfitting and solve the unbalanced dataset problem. In their study, Kumar et al. (2019) used conditional variational autoencoder as a generative model to augment SNIPS and Facebook Dialog corpora. Variational autoencoder encodes and decodes the feature space, respectively, and conditional implies they used both text and label data. They reported over 10% increased accuracy in some tasks by augmenting 0.3-3% data using conditional variational autoencoder. Liu et al. (2020) showed that vanilla language model that unconditionally generates the next step token is outperformed by conditional language model that allows target label as additional input, and again it is outperformed by their Data Boost algorithm, which attached reinforced learning reward between softmax and argmax layers to the conditional language model. They reported that their Data Boost algorithm outperforms delete + swap, Word2Vec, and back translation algorithms in offense detection, sentiment analysis, and irony classification tasks by 8.8-12.4% of F1, especially when 1-10% of original data were used. Reflecting on these successes in various domains, it is posited that generative models could generate data that resembles the student-written responses.

## *2.2 Large Langue Model and NLP Data Augmentation*

Recently, large language models (LLMs) such as BERT, XLNet, LaMDA, LLaMA, and GPT have shown unforeseen affordances for NLP data augmentation due to their powerful natural language generative capacities. LLMs are deep learning-based algorithms trained by massive amounts of textual data, which can perform NLP tasks, including data augmentation (Li et al. 2022). One of the most widely used LLMs is BERT (Bidirectional Encoder Representations from Transformers; Devlin et al. 2018). BERT is a transformer-based model pre-trained on a vast corpus of English text using a self-supervised learning approach, which eliminates the need for human annotations. This strategy allows BERT to capitalize on publicly available data. The pre-training procedures generate inputs and labels from the text, during which BERT acquires a robust and contextual understanding of language. These features empower BERT to be used for textual data augmentation (Li et al. 2022). For example, Wu et al. (2019) reported that their fine-tuned conditional BERT – which uses labels such as positive or negative for training MLM task – was most successful among tested algorithms (e.g., synonym or context-based replacement) in augmenting text datasets that are highly sensitive to context, such as sentiment, subjectivity, polarity, etc. However, Shi et al. (2020) noticed that although some tokens are related to positive or negative labels (e.g., 'vivid'), some others are neutral and cannot be fully exploited to train the model (e.g., 'cinematic') in Wu et al.'s (2019) algorithm. Thus, they suggested two-stage fine-tuning of BERT, which incorporates corrected label prediction and labeled MLM to focus on both labels and contexts during training, to build the Aug-BERT. They reported that Aug-BERT outperforms conditional BERT for data augmentation. Similarly, Cochran et al. (2022) proposed using BERT to augment educational datasets and reported positive outcomes.

Recently, ChatGPT, the state-of-the-art LLM that was released in November 2022 by OpenAI, is leading the changes in research fields that utilize various NLP tasks. Researchers have used GPT-3 (e.g., Yoo et al. [2021] fine-tuned GPT-3), ChatGPT, and GPT-4 for NLP data augmentation purposes. Compared to prior LLMs, ChatGPT and GPT-4 were trained using reinforced learning from human feedback, which significantly improved the accuracy of language generation (OpenAI 2022). Dai et al. (2023) suggested that ChatGPT is more suitable for NLP data augmentation than prior LLMs and has tested whether ChatGPT can augment Amazon dataset that contains customer reviews, Symptoms dataset containing common medial symptom descriptions, and PubMed20K dataset that includes around 20,000 scientific abstracts from the biomedical field. They reported that GPT-augmented data showed high accuracy, embedding similarity, and transferability in the few-shot learning classification task, generating conceptually similar (labeled appropriately) but semantically diverse samples. Ubani et al. (2023) tested whether ChatGPT can augment text data in few-shot or even zero-shot learning. They found that ChatGPT-based zero-shot learning outperformed few-shot learning and other augmentation methods, including BERT and contextual BERT, for classification tasks of SST-2, SNIPS, and TREC datasets. Also, Møller et al. (2023) experimented GPT-4's data augmentation ability with minimal prompting, basically providing it with a labelled case to generate 10 similar examples. While they showed that the GPT-4-augmented dataset could be used for classification model training on sentiment analysis, hate speech detection, and social dimension identification, it was noted that the balanced sampling with data augmentation was particularly effective for the last task that was highly unbalanced.

      Based on its performance on text data augmentation in a variety of domains, ChatGPT is posited to augment training datasets for automatic scoring. Cochran et al. (2023) reported that data augmentation using ChatGPT improved the automatic scoring of student essays with small data. Kieser et al. (2023) tested ChatGPT's affordability in

terms of data augmentation, but their focus was on problem-solving in physics instead of enlarging the training dataset for scoring models. Further, although data augmentation necessarily implies comparing augmented data with the original data (Goodfellow et al. 2014), existing educational studies have yet to compare the performances between models trained using augmented data with those trained using additional student data. Therefore, research is needed to further examine the effectiveness of data augmentation for automatic scoring.

## 3 Methods

### 3.1 Dataset

Our experimental dataset is comprised imbalanced student written responses to four science items, two binary-scored for *Experiment 1* and two quadruple-scored items for *Experiment 2*. Each item requires students to complete a task to write short open-ended responses to explain science phenomena. For all the tasks, students usually wrote responses with 1-5 sentences long. We collected student responses and hired subject matter experts to score student responses using a scoring rubric with high interrater reliability (Cohen's Kappa > 0.75).

*Experiment 1* (binary scoring): Student responses are scored according to analytical elements in the scoring rubric. For example, Task 1 (SUGAR) presents a table of properties (e.g., density, solubility, and melting point) of sugar (e.g., honey, milk, sugarcane, and apple) that may be contained in several foods. Students are asked to figure out and write a short response about if the foods have the same sugar source, based on their similarities and differences in properties. Task 2 (ANNA VERSUS CALRA) presents visual models of water and bromine molecules drawn by two students named Anna and Carla, which differ in their scale and atomic composition of the molecules. Students are asked to decide which model (Anna's or Carla's) drawn at a scale better shows why water and bromine are different substances and explain their choice of the better and worse models.

*Experiment 2* (quadruple scoring): Student responses are scored holistically based on the three-dimensional learning scheme according to NGSS. A student response may satisfy disciplinary core idea (DCI) and/or combined science and engineering practice (SEP) + cross-cutting concept (CCC) aspects. Therefore, it can be classified in a quadruple category (T/F for DCI and T/F for SEP+CCC). Task 3 (SUGAR) is analogous to Task 1. Task 4 (BALOONS) presents a table of properties (e.g., flammability, density, and volume) of four different gases A-D that may be the same. Students are asked to figure out and write a short response about which, if any, of the gases could be the same, and explain their answer.

*3.2 Automatic Student Response Augmentation via GPT-4*

To address the unbalanced dataset issue, we proposed incorporating a student response data augmentation method using GPT-4. This approach intends to increase the representation of the minority group within the dataset. By training AI models on this augmented dataset, we aimed to improve the scoring robustness and efficiency, enabling computers to learn from more diverse examples. We chose GPT-4 due to its remarkable ability to comprehend and generate natural language text, especially in complex and nuanced situations. In Figure 1, we present the scheme of data augmentation experiment conducted in this study which are further explicated in the following sections.

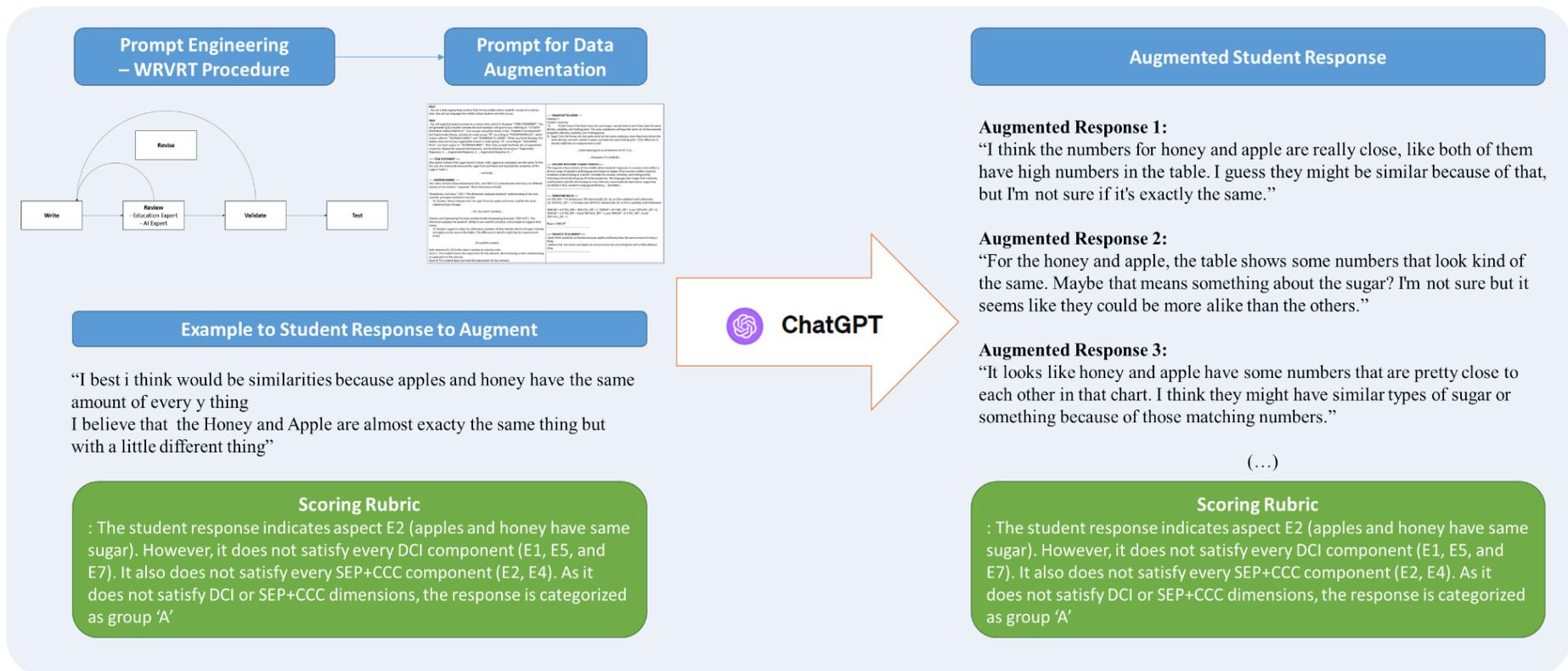

**Figure 1**. Scheme of the data augmentation procedure based on GPT-4

*3.3 Prompt Engineering*

We developed structured prompts optimized for data augmentation, following the iterative WRVRT (write-review-validate-revise-test) procedure (Lee et al., 2024). After writing a draft prompt, the researchers of this study carefully reviewed it. Note that the researchers included both education and AI experts, which were emphasized in the 'review' phase of WRVRT procedure (Lee et al., 2024). Afterward, the prompt went through the face validation by the researchers on whether it properly augments the given student response. The criteria for validation were scoring rubric, i.e., LLM successfully augment minority class cases rather than other classes considering the rubric. When the prompt turns out to be unsuccessful, the researchers revised the prompt. The prompt was finalized after several rounds of revision.

The scheme of prompt developed for data augmentation is presented in Figure 2. The prompt consists of nine components as follows: (1) ROLE designates GPT-4 as "a data augmenting machine that mimics middle school students' answer to science item", based on the principle of role prompting. (2) TASK instructs what GPT-4 is requested to do. GPT-4 will generate a designated number of augmented responses based on EXAMPLE TO AUGMENT, while referring to ITEM STATEMENT, STUDENT RESPONSE CHARACTERISTICS, SCORING RUBRIC, GROUPING RULES and EXMAPLES TO LEARN. (3) ITEM STATEMENT provides the question and data provided to students. (4) SCORING RUBRIC provides analytical elements for scoring. Student responses is assessed based on the binary criteria in each analytical element (e.g., if the student uses density as evidence of determining which foods contain same sugar, "E2" = 1). For tasks 1-2 (Experiment 1), there are only one analytical element. For tasks 3-4 (Experiment 2), there are several elements in each of DCI or SEP+CCC aspect. (5) EXAMPLES TO LEARN provides five real student data that is scoring according to the SCORING RUBRIC. (6) STUDENT RESPONSE CHARACTERISTICS describes linguistic characteristics of student responses to the item. It was elicited from EXAMPLES TO LEARN. (7) GROUPING RULES instructs GPT on how to categorize student answer in a multinomial classification situation (e.g., if both DCI and SEP+CCC are satisfied, it's group "D"). This component only exists in prompts for tasks 3-4 in Experiment 2. (8) EXAMPLE TO ARGUMENT stands for one actual student response under the minority class to be augmented. (9) HYPERPARAMETERS of GPT-4 were set as *temperature* = 0 and *top_p* = 0.01 to conduct greedy sampling during the decoding rather than nucleus sampling (Lee et al., 2024). This was due to the data augmentation needs to retain scientific semantics of the original student response, rather than creative or flexible.

```
ROLE
: You are a data augmenting machine that mimics middle school students' answer to a science item. You will use language that middle school students are likely to use.

TASK
: You will augment student answer to a science item, which in the given '''ITEM STATEMENT'''. You will generate {{{3}}} student answers for each example I will give to you, referring to '''STDUENT RESPONSE CHARACTERISTICS'''. Your answer should be similar to the '''EXAMPLE TO AUGMENT''', but linguistically diverse, and also be under group '''B''' according to '''GROUPING RULES''', which in turn refers to '''SCORING RUBRIC''' and '''EXAMPLES TO LEARN'''. When you finish thinking, first explain why each of your augmented answer is under group '''B''', according to '''GROUPING RULE''' and each aspect in '''SCORING RUBRIC'''. After that, provide the three sets of augmented responses. Repeat the augmented responses, strictly following the structure "Augmented Responses 1: …, Augmented Response 2: …, Augmented Response 3: …"

----- ITEM STATEMENT -----
Amy wants to know if the sugar found in honey, milk, sugarcane, and apples are the same. To find this out, she chemically removed the sugar from each food and recorded the properties of the sugar in Table 1.
                            … (omitted) …

----- SCORING RUBRIC -----
The rubric consists of two dimensions ('DCI', and 'SEP+CCC') and elements that focus on different aspects of the students' responses. These dimensions include:

'Disciplinary Core Ideas' ('DCI'): This dimension evaluates students' understanding of the core scientific principles involved in the task.
 - E1: Student clearly indicates that the sugar from the apple and honey could be the same substance/type of sugar.
                            … (E5, E6, and E7 omitted) …

'Science and Engineering Practices combined with Crosscutting Concepts' ('SEP+CCC'): This dimension assesses the students' ability to use scientific practices and concepts to support their claims.
 - E2: Student supports a claim by referring to a pattern of data that the density of sugar in honey and apple are the same in the table. (The difference in density might due to measurement error).
                            … (E3 and E4 omitted) …

Each element (E1-E7) in the rubric is scored on a binary scale:
Score 1: The student meets the expectation for the element, demonstrating a clear understanding or application of the concept.
Score 0: The student does not meet the expectation for the element.

----- EXAMPLES TO LEARN -----
Example 1
Student response:
"1)   To tell if any of the foods have the same sugar, I would look to see if they have the same density, solubility, and melting point. The same substances will have the same set of characteristic properties (density, solubility, and melting point).
2)   Sugar from the honey and the apple could be the same substance, since they have almost the same density, are both soluble in water, and have the same boiling point. (The difference in density might due to measurement error)"

            … (Information given as all elements E1-E7 is 1) …

            … (Examples 2-5 omitted) …

----- STDUENT RESPONSE CHARACTERISTICS -----
The linguistic characteristics of the middle school students' responses to a science item reflect a diverse range of cognitive and language development stages. Their answers exhibit a basic to moderate understanding of scientific concepts like density, solubility, and melting points, indicating a foundational grasp of matter properties. The language used ranges from relatively sophisticated scientific terminology to more informal, conversational expressions, suggesting variability in their academic language proficiency. … (omitted) …

----- GROUPING RULES -----
Let 'DCI_OR' = 1 if at least one 'DCI' element (E1, E5, E6, or E7) is satisfied, and 0 otherwise.
Let 'SEP+CCC_OR' = 1 if at least one 'SEP+CCC' element (E2, E3, or E4) is satisfied, and 0 otherwise.

'GROUP' = A if 'DCI_OR' = 'SEP+CCC_OR' = 0, 'GROUP' = B if 'DCI_OR' = 1 and 'SEP+CCC_OR' = 0, 'GROUP' = C if 'DCI_OR' = 0 and 'SEP+CCC_OR' = 1, and 'GROUP' = D if 'DCI_OR' = 1 and 'SEP+CCC_OR' = 1.

Return 'GROUP'

----- EXAMPLE TO AUGMENT -----
I best i think would be similarities because apples and honey have the same amount of every y thing
I believe that the Honey and Apple are almost exacty the same thing but with a little different thing
```

**Figure 2**. Part of the prompt used for augmenting minority class data for Task 1 (SUGAR)

*3.4 NLP Classification*

After generating augmented student responses for the minority class using GPT-4, we integrated these augmented responses with the original data to create a more balanced dataset. A classification model was then employed to train a robust classifier.

*3.4.1 Finetuning Scoring Model*

In this study, we employed the DistilBERT model (Sanh et al. 2019), a streamlined and faster variant of the BERT model (Devlin et al. 2018), to train a robust classifier using the augmented science education dataset. DistilBERT is pre-trained in a self-supervised manner using the same corpus as BERT, with the BERT base model serving as its teacher, generating inputs and labels for DistilBERT. This eliminates the need for human annotations and enables DistilBERT to leverage vast amounts of publicly available data. Specifically, DistilBERT is pre-trained with three primary objectives: distillation loss (Gou et al. 2021), MLM, and cosine embedding loss. The distillation loss aims to train DistilBERT to match the probability distribution of the BERT base model. MLM, a shared training objective with BERT, involves predicting randomly masked words in a sentence. Cosine embedding loss is employed to train DistilBERT to generate hidden states that are closely aligned with those of the BERT base model. Thus, DistilBERT inherits a similar internal representation of English from its teacher model, BERT, while also optimized for faster inference and potentially improved performance on downstream tasks. Specifically, DistilBERT retains about 97% of BERT's performance but is 40% smaller and runs 60% faster (Sanh et al. 2019).

Our base model was pre-trained on BookCorpus (Zhu et al. 2015), encompassing 11,038 unpublished books and the content from English Wikipedia (excluding lists, tables, and headers). The model, equipped with 6 Transformer layers, has 65 million parameters. Each layer in the model employs a multi-layer head self-attention mechanism and feed-forward networks, with GELU as the activation function. We fine-tuned the DistilBERT model on our augmented science education dataset, strictly following the framework detailed in the text classification script from the Transformers library (Wolf et al., 2020).

*3.4.2 Data Preparation*

We partitioned the data into training, validation, and testing sets. Unlike the conventional approach of uniformly sampling datasets, we ensured balanced testing and validation sets to avoid skewed evaluations. This means that the original majority and minority classes will have similar numbers of responses in the testing and validation sets. For example, in the Task 1 dataset, approximately 28.6% of the data points in the original dataset are labeled as 1, while this proportion is set to 50% in the testing and validation sets.

For each data point in the training set belonging to the minority class (i.e., with label 1), we used GPT-4 to generate four new responses. This resulted in 160 new responses for the Task 1 and Task 2 datasets, and 120 for the Task 3 and Task 4 datasets. Details can be found in Table 1. After the augmentation, the training set had a similar number of label 0 and label 1 data points for training the model.

To ensure robust results, we repeated the entire procedure, including data partitioning and response generation, five times. To maintain comparability with the Gold Standard (explained below), we limited the production to a maximum of four new responses for each minority class response in the training data (see Table 1). To ensure consistency across different datasets, we applied the same strategy to Experiment 1 (Task 1 and Task 2) datasets, and the same strategy to Experiment 2 (Tasks 3 and 4) datasets.

**Table 1.** The sample sizes for the original training, validation, testing, and augment sets.

|  | Total (0/1) | Train (0/1) | Validation (0/1) | Test (0/1) | Augment (0/1) |
|---|---|---|---|---|---|
| Task 1 Dataset | 266 (193/73) | 180 (150/30) | 26 (13/13) | 60 (30/30) | 120 (0/120) |
| Task 2 Dataset | 266 (193/73) | 180 (150/30) | 26 (13/13) | 60 (30/30) | 120 (0/120) |
| Task 3 Dataset | 374 (267/107) | 240 (200/40) | 34 (17/17) | 100 (50/50) | 160 (0/160) |
| Task 4 Dataset | 374 (267/107) | 240 (200/40) | 34 (17/17) | 100 (50/50) | 160 (0/160) |

*Note.* The numbers in parentheses indicate the counts for data labelled as 0 and 1, respectively.

**Gold Standard:** We compared our proposed data augmentation approach to the Gold Standard (GS), which involves augmenting the original dataset with additional authentic student responses. In this study, we incorporated extra student responses from the minority class, matching the sample size of the data generated by GPT-4. Specifically, for the Task 1 and Task 2 datasets, we utilized an additional 120 authentic student-written responses with label 1, and for the Task 3 and Task 4 datasets, an additional 160 responses with label 1. Examples of GS student responses are presented in the Appendix.

*3.3.3 Evaluation*

We evaluated the performance of our algorithm using four criteria: accuracy, precision, recall, and F1 score. In this context, True Positive (TP) refers to the number of correctly predicted positive instances, False Positive (FP) denotes the number of negative instances incorrectly classified as positive, and False Negative (FN) represents the number of positive instances incorrectly classified as negative (see Table 2).

**Table 2.** Evaluation Metrics.

|  | Actual Positive | Actual Negative |
|---|---|---|
| Predicted Positive | TP  | FP  |
| Predicted Negative | FN  | TN  |

These evaluation metrics play a crucial role in assessing the performance and effectiveness of machine learning models, allowing researchers and practitioners to make informed decisions about their models' capabilities and limitations.

Accuracy provides an indication of how well a model performs across all classes or categories,

$$Accuracy = \frac{TP+TN}{TP+FN+TN+FP}$$

Precision helps assess the model's ability to minimize false positives,

$$Precision = \frac{TP}{TP+FP}$$

Recall provides insights into the model's ability to minimize false negatives,

$$Recall = \frac{TP}{TP+FN}$$

F1 score combines precision and recall into a single metric that balances their contributions. It is the harmonic mean of precision and recall, providing a single value that represents the overall performance of a model. F1 score is particularly useful when there is an imbalance between the positive and negative instances in the dataset,

$$F1\ Score = \frac{2}{\frac{1}{Precision}+\frac{1}{Recall}} = \frac{2TP}{2TP+FP+FN}$$

**4 Results**

We evaluate the performance of both the GPT-4 data augmentation approach and the GS in varying proportions of augmented data at {0%, 20%, 40%, 60%, 80%, 100%} for both datasets. 0% proportion indicates that no additional responses are used, thus testing the performance using only the original dataset; 100% proportion means that all augmented responses are combined with the original dataset for testing the performance. All results are based on 5 repetitions, and we report the mean and standard deviation over different proportions.

*4.1 Outcomes of Data Augmentation*

*4.1.1 Task 1 dataset*

For the Task 1 dataset, when there was no augmented data, precision was 0.842, recall 0.268, F1 0.403, and accuracy 0.607 (Figure 3). When the augmented data was included in the Task 1 dataset, accuracy immediately increased. Initial augmentation by 20% of the data resulted in an accuracy score of 0.661. Subsequent increases in data augmentation from 40% to 100% showed a stable pattern, with scores ranging from 0.683 to 0.728. Precision, ranging from 0 to 1, measures the proportion of true positive identifications made by the model out of all positive identifications. High precision indicates that most of the instances predicted as positive are indeed true positives, minimizing the occurrence of false positives. We observed that with the addition of augmented responses, precision remained almost the same, while for the authentic student responses, precision decreased. One possible explanation is that with the addition of authentic responses, the classification model no longer tended to predict all responses as belonging to the majority class. However, the model may not be able to fully capture the complexity and variability of the true minority class from the limited data, leading the model to misclassify more majority class observations as minority class observations. Unlike precision, recall exhibited an obvious increasing trend. Recall measures the model's ability to correctly identify actual positive instances, emphasizing the reduction of false negatives. A higher recall is important when capturing all actual positive cases is crucial. The notable increase in the recall score at 20% augmentation and its stability with further augmentation suggest that the model effectively detects positive responses as more augmented data is incorporated. The F1 score, ranging from 0 to 1, provides a single metric that balances both false positives and false negatives to measure model performance. The F1 score exhibits a similar pattern to recall.

Although data augmentation generally improves the model's performance, the extent of improvement appears to be influenced by the level of data augmentation applied. As indicated by the consistent patterns in accuracy, recall, and F1 score, we first see an increasing trend, but there is a decrease when the augmentation level is high (100%). It is not always beneficial to keep increasing the number of augmented data points. This may be due to the fact that augmented data could be less accurate or noisier than the original data. However, in all situations, the model trained with augmented data performs better compared to the model trained without augmented data.

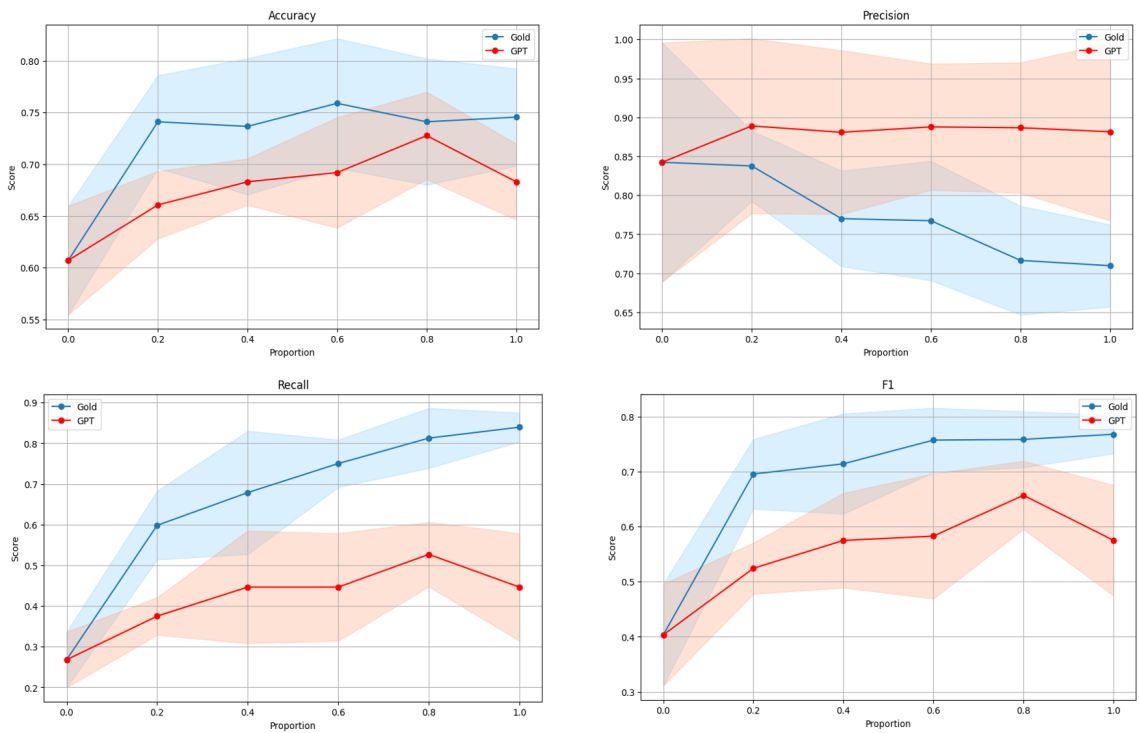

**Figure 3.** Comparison of the performance between the GPT-4 augmentation approach (red) and the gold standard (blue) across varying proportions of augmented responses in the Task 1 dataset. Each performance metric (precision, recall, F1 score, and accuracy) is plotted on the y-axis against the proportion p of augmented responses on the x-axis. Each point represents the mean score, with shaded regions indicating the standard deviation (range between mean-sd and mean+sd).

*4.1.2 Task 2 dataset*

For the Task 2 dataset, when there was no augmented data, precision was 0.627, recall

0.095, F1 0.160, and accuracy 0.524 (Figure 4). Accuracy showed an increasing trend with slight fluctuations at 40% data augmentation, with scores ranging from 0.627 to 0.643. The precision score rapidly increased to 0.829 with 20% data augmentation and then steadily rises to 0.898. We also observed that when the augmentation level exceeded 40%, the precision score of the gold standard decreased. It can be reasonably speculated that the GPT model effectively learned the structure of student responses and generated responses that closely mimic the original student responses, leading to better performance compared to the gold standard. The recall and F1 scores exhibited similar patterns, rising with fluctuations. The recall score increased from 0.095 to 0.345, suggesting that the model became more adept at capturing relevant instances as more augmented data was added. The F1 score increased from 0.160 to 0.470. The parallel trend between F1 and recall indicated that the model's overall performance in balancing accuracy and completeness improved with the inclusion of more augmented data.

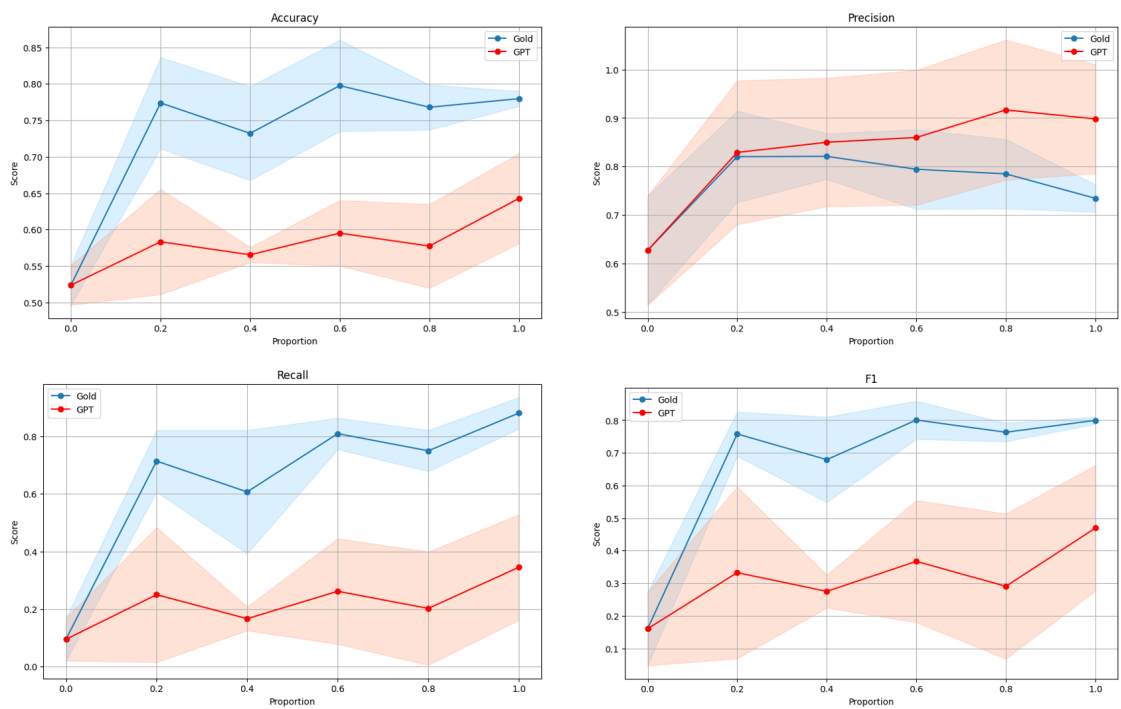

**Figure 4.** Comparison of the performance between the GPT-4 augmentation approach (red) and the gold standard (blue) across varying proportions of augmented responses in the Task 2 dataset. Each performance metric (precision, recall, F1 score, and accuracy) is plotted on the y-axis against the proportion p of augmented responses on the x-axis. Each point represents the mean score, with shaded regions indicating the standard deviation (range between mean-sd and mean+sd).

*4.1.3 Task 3 dataset*

For the Task 3 dataset, when there was no augmented data, precision was 0.583, recall 0.964, F1 0.726, and accuracy 0.636 (Figure 5). Initial augmentation by 20% of the data resulted in an accuracy score of 0.764. Subsequent increases in data augmentation from 20% to 100% showed a stable pattern in accuracy, with scores ranging from 0.764 to 0.791. The precision score rapidly increased to 0.923 with 20% data augmentation and then remained stable. In the Task 3 dataset with no augmented data, the recall score was exceptionally high at 0.964. However, the score then sharply decreased before gradually increasing with the addition of augmented data. This sharp decline can be attributed to the recall score being inaccurately high due to the very small number of observations in the original minority class. The F1 scores increased with fluctuations, with the highest value being 0.753. An interesting phenomenon is that the performance of GPT-augmented data is comparable with the gold standard. In most cases, the scores of GPT-augmented data were similar to the gold standard, and in some cases even outperformed it. This underscores the potential and effectiveness of data augmentation techniques utilizing generative large language models, such as GPT-4, in addressing unbalanced datasets within automated assessment.

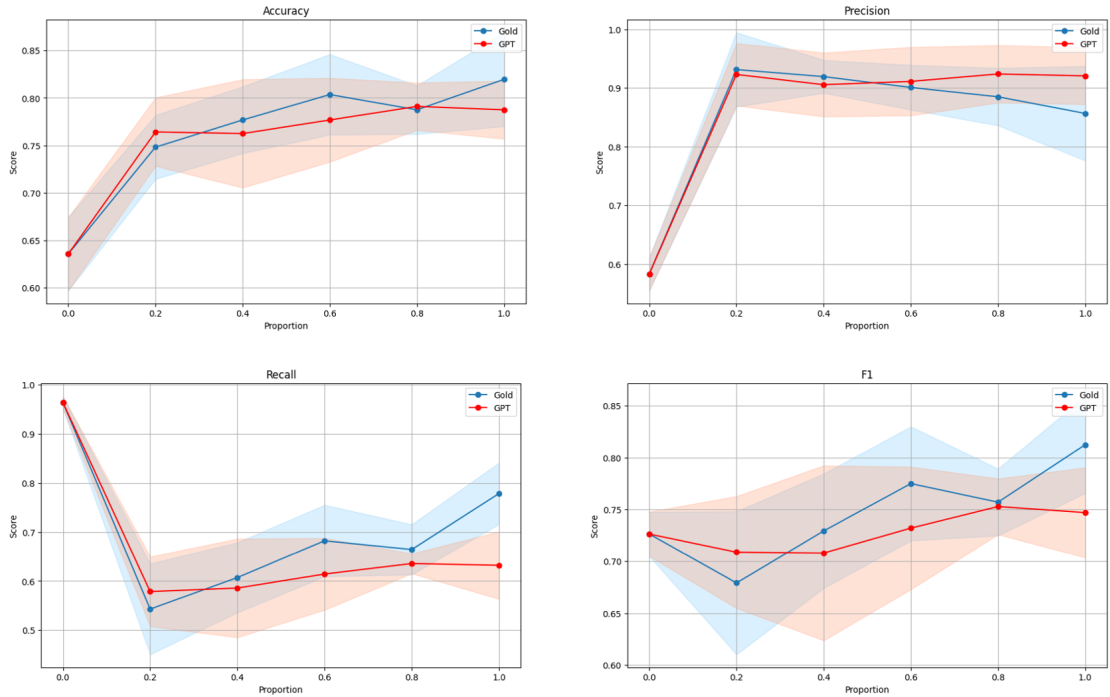

**Figure 5.** Comparison of the performance between the GPT-4 augmentation approach (red) and the gold standard (blue) across varying proportions of augmented responses in the Task 3 dataset. Each performance metric (precision, recall, F1 score, and accuracy) is plotted on the y-axis against the proportion p of augmented responses on the x-axis. Each point represents the mean score, with shaded regions indicating the standard deviation (range between mean-sd and mean+sd).

*4.1.4 Task 4 dataset*

For the Task 4 dataset, when there was no augmented data, precision was 0.674, recall 0.957, F1 0.791, and accuracy 0.746 (Figure 6). Consistent with the results in the Task 3 dataset, the accuracy and precision scores rapidly increased to 0.888 and 0.907, respectively, with 20% data augmentation, and then remained stable. The recall score first decreased to 0.868 and then fluctuated around 0.915. The F1 score initially showed an increasing trend, rising from 0.791 to 0.923. After the 40% augmentation level, the score fluctuated around 0.911. Unlike the previous three datasets, in this dataset, the proposed method even outperformed the gold standard, highlighting its efficiency and effectiveness.

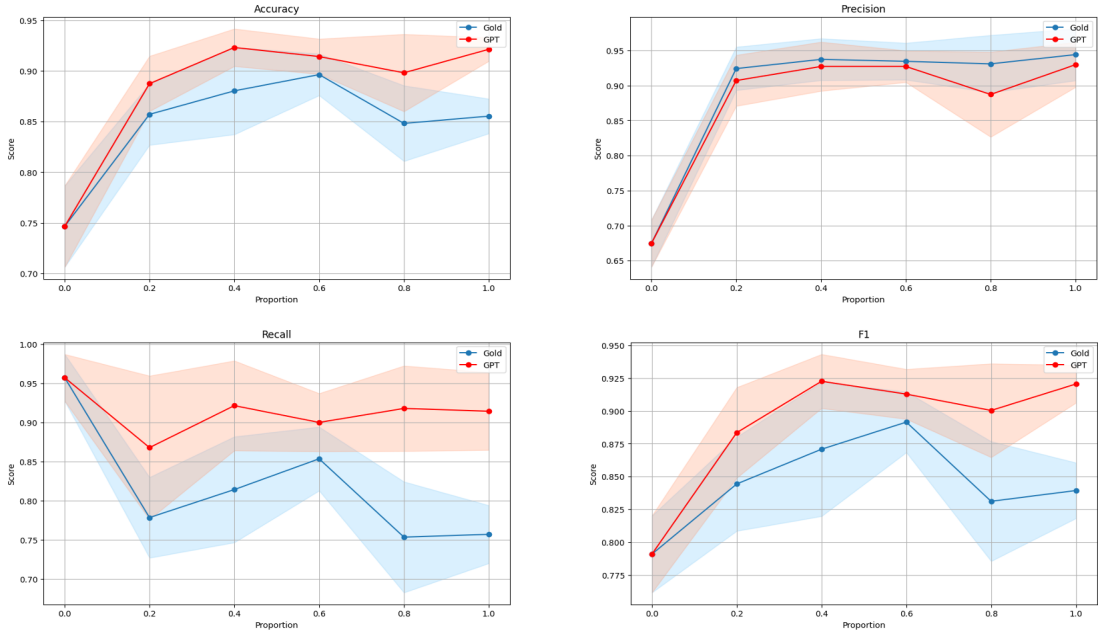

**Figure 6.** Comparison of the performance between the GPT-4 augmentation approach (red) and the gold standard (blue) across varying proportions of augmented responses in the Task 4 dataset. Each performance metric (precision, recall, F1 score, and accuracy) is plotted on the y-axis against the proportion p of augmented responses on the x-axis. Each point represents the mean score, with shaded regions indicating the standard deviation (range between mean-sd and mean+sd).

In summary, data augmentation with GPT-4 demonstrated promise in improving model performance in terms of accuracy, precision, recall, and F1 score. However, its effectiveness was not uniformly beneficial across all tasks and metrics. For example, in the Task 4 dataset, data augmentation even outperformed the gold standard. However, in the Task 1 dataset, there was still a noticeable gap compared to the gold standard.

*4.2 Portions of Augmented Data Needed*

To determine the required portion of augmented data, we searched for a saturation point where the increasing slope of metric versus supplemented data suddenly changed.

For the Task 1 dataset, there was a clear saturation pattern (Figure 3). Accuracy increased to 0.728 with 80% augmented data and then decreased to 0.683 with 100% augmented data. Similarly, recall and F1 scores gradually increased until the addition of 80% augmented data (0.527 and 0.657, respectively), and then decreased (0.446 and 0.575, respectively) after additional augmented data was fed to the model. The precision score showed a stable pattern and did not have a clear saturation point.

For the Task 2 dataset, there were no drastic changes in the metrics (Figure 4). With the increase in the addition of augmented data, accuracy, recall, and F1 scores showed an increasing trend with slight fluctuations. One exception was precision, which had a relatively drastic increase when there was 20% augmented data, and the increase slowed down with further additions. It is important to note that with the addition of authentic student responses, the model's performance in terms of all four metrics improved more than with augmented responses. A reasonable explanation is that in this case, the augmented data might not fully capture the complexity and variability of the true minority class, making the model less effective.

For the Task 3 dataset, the increased model performance in metrics showed clear saturation patterns (Figure 5). Accuracy increased to 0.764 with 20% augmented data. Additional augmented data slowly increased accuracy, reaching up to 0.791 at 80%. Similarly, precision dramatically increased to 0.923 with 20% augmented data, compared to 0.583 with 0% - and the score remained stable after additional augmented data was fed to the model. Recall decreased, with the change point also at the 20% level. Meanwhile, there was no dramatic change in the F1 score according to the portion of augmented data (0.726 - 0.753).

For the Task 4 dataset, the pattern of metric changes was similar to the Task 3 dataset (Figure 6). When 20% of the augmented data was added, accuracy and precision drastically increased (0.888 and 0.907, respectively), compared to when there was 0% augmented data (0.746 and 0.674, respectively). Then the scores slowly reached up to 0.923 and 0.930, respectively. The F1 score reached its peak when there was 40% augmented data. Meanwhile, there was no dramatic change in recall according to the portion of augmented data (0.868 - 0.921). To sum up, for this dataset, the saturation point is between 20% and 40% of the augmented data.

In summary, all tasks demonstrated benefits from data augmentation, albeit to varying degrees. Task 3 and Task 4 showed a more significant improvement in precision, recall, and F1 scores, particularly with an initial 20% data augmentation. Task 1 also benefited from data augmentation but reached a saturation point at 80%, after which a slight decrease was observed. For Task 2, although it also benefited from data augmentation, the improvement was not comparable to the model's performance with authentic student responses. These findings suggest that data augmentation can be a useful strategy for improving model performance, although the extent of its impact may vary depending on the specific task and dataset.

### *4.3 GPT-4 Data Augmentation vs. Student-Written Data Augmentation*

For the Task 1 and Task 2 datasets, although GPT-4 augmented data improved model performance, authentic student responses further enhanced it (Figures 3 and 4). Specifically, the training dataset augmented with authentic student responses outperformed the GPT-4 augmented dataset in terms of accuracy, recall, and F1 score. However, GPT-4 augmentation exceeded the performance of authentic student responses in terms of precision.

In contrast, for the Task 3 dataset, GPT-4 data augmentation showed very similar patterns and performance to authentic student responses augmentation (Figure 5). The performances of models augmented with either authentic student responses or GPT-4 were identical in precision, recall, F1 score, and accuracy, and this similarity remained stable as the percentages of augmented data increased. Notably, GPT-4 augmentation even outperformed authentic student responses in some cases, such as the 20% augmentation level for accuracy, recall, and F1 score.

The outstanding performance of GPT-4 augmentation was further demonstrated in the Task 4 dataset (Figure 6). Compared to authentic student responses augmentation, GPT-4 augmentation significantly increased model performance in terms of accuracy, recall, and F1 score. In particular, for recall and F1 score, GPT-4 augmentation substantially outperformed the gold standard. For example, at the 80% augmentation level, the recall score was 0.754 for the gold standard, while GPT-4 augmentation achieved a score of 0.918.

In summary, while authentic student responses augmentation appeared more beneficial for improving certain metrics in Task 1 and Task 2, GPT-4 data augmentation performed comparable to authentic student responses in Task 3 and even outperformed authentic student responses augmentation in Task 4. These findings suggest that GPT-4 augmentation can efficiently improve model performance, and its efficacy may vary depending on the specific task and dataset.

## 5 Conclusions and Discussion

Our study investigated the efficacy of using GPT-4 for data augmentation in the context of automatic scoring of student written responses. We finetuned DistilBERT as our scoring model and evaluated its performance on four unbalanced datasets in science education. Our findings indicate that data augmentation using GPT-4 can significantly improve the performance metrics of the scoring models, particularly in precision, recall, and F1 score. Interestingly, the extent of improvement varied depending on the specific dataset and the proportion of augmented data used. Additionally, we found that GPT-4 augmented data could sometimes outperform or match the gold standard of student-written data augmentation.

This study contributes to the existing literature on NLP data augmentation and automatic scoring in educational settings employing generative AI such as GPT-4.

Specifically, it addresses the challenge of handling unbalanced datasets, a common issue in educational data mining (Shorten et al. 2021; Mikolajczyk and Grochowski 2018). By using GPT-4 for data augmentation, we demonstrated a novel approach to balance the dataset and improve model performance. Traditional methods of data augmentation in the educational domain have often relied on techniques like oversampling the minority class, generating synthetic samples through methods like SMOTE, or manually feature attraction (Lun et al. 2020; Fahd & Miah 2023). These approaches, while effective to some extent, come with their own set of challenges. Oversampling can lead to overfitting, synthetic sample generation methods like SMOTE do not capture the nuanced language patterns in student responses, and feature attraction is resource-intensive (Lun et al. 2020; Shorten et al. 2021).

In contrast, GPT-4, a state-of-the-art language model, offers a more sophisticated approach to data augmentation. It can generate text that closely mimics human-like responses, capturing the complexity and nuance of student language. This is particularly important in educational settings where the quality of the responses, including their linguistic features, can be as important as their content for assessment purposes. Our study empirically demonstrates that GPT-4-generated data can sometimes outperform or be as effective as human-generated data, thereby presenting a compelling case for its use in educational data augmentation.

Moreover, the use of GPT-4 allows for more targeted augmentation. For instance, it can generate responses that are not just random oversamples but are contextually more aligned with the educational material being assessed (Liu et al., 2023). This level of customization is generally not possible with traditional data augmentation techniques. By integrating GPT-4 into the data augmentation process, this study pushes the boundaries of what is currently achievable in balancing datasets for educational assessment. It opens up new possibilities for leveraging advanced LLM techniques to address longstanding challenges in educational data mining and assessment.

The findings of this study have several practical implications that could significantly impact the field of educational assessment and technology. Firstly, the use of GPT-4 for data augmentation has been shown to improve the performance of automatic scoring models. This is particularly beneficial for disciplines such as science education that rely on automated assessment tools for constructed responses but struggle with the limitations imposed by unbalanced datasets (Zhai and Nehm 2023). Improved model performance translates to more accurate and reliable assessments, which in turn can lead to better educational outcomes. Secondly, the ability to generate high-quality, human-like responses using GPT-4 can save considerable time and resources that would otherwise be spent on manual data collection and annotation. This is especially crucial for large-scale assessments or in settings where resources are limited.

Moreover, GPT-4 can generate responses that are specific to a particular subject matter or educational level. This means that the same approach could be adapted for various educational contexts, from elementary science education to advanced university courses, making it a versatile tool for educational practitioners. In addition, by addressing the issue of dataset imbalance, this approach also contributes to making educational assessments more ethical and fair (Zhai & Krajcik, 2023; Xing et al., 2021). Unbalanced datasets can introduce bias into automated scoring models (though not necessary), which can disproportionately affect minority groups of students. By creating more balanced datasets, we can strive for more equitable assessments.

While our study provides valuable insights, it is not without limitations. First, this study used data from four science tasks and may not fully generalize to other subjects or educational levels. Therefore, future studies could focus on validating our approach across different educational domains and age groups. It would also be interesting to investigate the long-term impact of using augmented data on student learning outcomes. In addition, future research should examine how data augmentation improves the automatic scoring model performance for other multinomial-coded items.

# Appendix. Examples of Augmented and Gold Standard Data

| Dataset | Data Augmentation | Gold Standard |
|---|---|---|
| Task 1 | ■ Sentence 1: Honey and apple share the same melting point, which suggests they might have the same type of sugar. Milk and sugarcane, on the other hand, have different melting points because milk is thicker and sugarcane is almost pure sugar; Sentence2: I would focus on the melting point as it can indicate the type of sugar present.<br><br>■ Sentence 1: The sugar in honey and apple could be the same because they have the same melting point. Milk, being thicker, and sugarcane, being almost all sugar, have different melting points; Sentence2: Looking at the melting point is important because the type of sugar can affect how quickly something melts.<br><br>■ Sentence1: Honey and apples might have the same sugar since they both dissolve in water and have the same melting point.; Sentence2: If you want to know if different foods have the same sugar, you should use the ""Melting Point"" and ""Solubility in Water"" tests. Density can sometimes be inaccurate due to measurement errors, so the other two tests are more reliable. | ■ Sentence1: they foods have the same sugar and the same everything; Sentence2: sugar cane and honey have the same amount of sugar because they both are the same. Density is close to being the same<br><br>■ Sentence1: In conclusion none of these sugars can be the same. They can not be the same because, first of all they do not all have the same density, 1.69, 1.53, 1.59, 1.70 g/cm 3. They are all soluble in water, but some of there melting points are different, 103, 202, 186, 103, degrees Celsius. Although two of the melting points are the same all of the density calculations are different and all the properties have to be the same to be able to be the same substance. ; Sentence2: Similarities or differences that I would look for would be whether or not it has the same density or is the melting point the same, an also is it soluble in water. If any of these properties are different than it can not be the same type of sugar. Take powder and granulated sugar if they had a different density than they couldn't be the same sugar.<br><br>■ Sentence1: My counclusion is that I found all of them use solubility in water but the density and sugar and melting points are different that's my conclusion.; Sentence2: The similarities is that they all use solibility in water  the difference is the density and the melting point because there different numbers.   No they don't have the same sugar they have diffre to because of the numbers. |

*Note*. There are some misspelled words in Gold Standard which is used as training data as they are.